\documentclass[conference]{IEEEtran}
\usepackage{times}

\usepackage[numbers]{natbib}
\usepackage{multicol}
\usepackage[bookmarks=true]{hyperref} 

\IEEEoverridecommandlockouts
\usepackage{nccmath,amsmath,amssymb,amsfonts}
\usepackage{algorithm}
\usepackage{algorithmic}
\usepackage{graphicx}
\usepackage{textcomp}
\usepackage{xcolor}
\usepackage{comment}
\usepackage{lipsum}                     % Dummytext
\usepackage{xargs}                      % Use more than one optional parameter in a new commands
\usepackage[pdftex,dvipsnames]{xcolor}  % Coloured text etc.
\usepackage[colorinlistoftodos,prependcaption,textsize=tiny]{todonotes}
\usepackage{array}
\usepackage{caption}
\usepackage{float}
\usepackage{multirow}
\def\BibTeX{{\rm B\kern-.05em{\sc i\kern-.025em b}\kern-.08em
    T\kern-.1667em\lower.7ex\hbox{E}\kern-.125emX}}
\usepackage{tabularx}

\usepackage{xspace}
\usepackage{hyperref}

\newcommand{\eg}{e.g.}
\newcommand{\etc}{etc.}
\newcommand{\ie}{i.e.}

\newcommand{\hidrive}{Hi-Drive\xspace}
\newcommand{\algname}{Vec-QMDP\xspace}

\newcommand{\secref}[1]{Section~\ref{#1}}
\newcommand{\algoref}[1]{Algorithm~\ref{#1}}
\renewcommand{\eqref}[1]{(\ref{#1})}
\newcommand{\tabref}[1]{Table~\ref{#1}}

\newcommand{\figref}[1]{Fig.~\ref{#1}}

\newcommand{\egocar}{ego-vehicle\xspace}

\usepackage{etoolbox}
\newtoggle{final}
\definecolor{darkred}{rgb}{0.6 0 0}
\definecolor{black}{rgb}{0 0 0}
\newcommand{\modified}[1]{\iftoggle{final}{#1}{{\color{black} #1}}}

\newcommand{\ci}[1]{{\scriptsize$\pm$#1}} % or smaller with \scriptsize

\usepackage{fontawesome}

\newcommand{\blackdown}{\textcolor{black}{$\downarrow$}}

\usepackage{graphicx}
\usepackage{caption}
\usepackage{subcaption}
\usepackage{float}
\usepackage{wrapfig}
\usepackage[export]{adjustbox}
\usepackage{xparse}
\usepackage{booktabs}
\usepackage{calc}
\usepackage{pgfkeys}

\makeatletter
\def\maxwidth#1{\ifdim\Gin@nat@width>#1 #1\else\Gin@nat@width\fi}
\makeatother

\usepackage{caption}

\NewDocumentCommand{\insertFigure}{O{ht!} O{0.8\linewidth} m m m O{}}{
    \begin{figure}[#1]
        \centering
        \includegraphics[width=#2, #6]{#3}
        \caption{#4}
        \label{#5}
    \end{figure}
}

\NewDocumentCommand{\wideFigure}{O{ht!} O{0.9\textwidth} m m m O{}}{
    \begin{figure*}[#1]
        \centering
        \includegraphics[width=#2, #6]{#3}
        \caption{#4}
        \label{#5}
    \end{figure*}
}

\newcommand{\sideBySideFigure}[7][ht!]{
    \begin{figure}[#1]
        \centering
        \begin{subfigure}[b]{0.49\columnwidth}
            \centering
            \includegraphics[width=\linewidth]{#4}
            \caption{#6}
            \label{fig:#3_a}
        \end{subfigure}
        \hfill
        \begin{subfigure}[b]{0.49\columnwidth}
            \centering
            \includegraphics[width=\linewidth]{#5}
            \caption{#7}
            \label{fig:#3_b}
        \end{subfigure}
        \caption{#2}
        \label{#3}
    \end{figure}
}

\pgfkeys{
 /fourgrid/.is family, /fourgrid,
 maincaption/.store in = \fMainCap,
 label/.store in       = \fLabel,
 imgA/.store in        = \fImgA,
 imgB/.store in        = \fImgB,
 imgC/.store in        = \fImgC,
 imgD/.store in        = \fImgD,
 capA/.store in        = \fCapA,
 capB/.store in        = \fCapB,
 capC/.store in        = \fCapC,
 capD/.store in        = \fCapD,
}

\graphicspath{{figures/}}

\begin{document}

\title{\algname: Vectorized POMDP Planning on CPUs for Real-Time Autonomous Driving \\

%\algname
}

% \author{Author Names Omitted for Anonymous Review. Paper-ID 78}

\author{
    \IEEEauthorblockN{
        Xuanjin Jin\IEEEauthorrefmark{1}\IEEEauthorrefmark{2}, 
        Yanxin Dong\IEEEauthorrefmark{1}, 
        Bin Sun\IEEEauthorrefmark{3},
        Huan Xu\IEEEauthorrefmark{3},
        Zhihui Hao\IEEEauthorrefmark{3},
        XianPeng Lang\IEEEauthorrefmark{3},
        Panpan Cai\IEEEauthorrefmark{1}\IEEEauthorrefmark{2}
    }

    \IEEEauthorblockA{
        \IEEEauthorrefmark{1}Shanghai Jiao Tong University, 
        \{xuanjin.jin, cai\_panpan\}@sjtu.edu.cn, yanxinn.d@gmail.com 
    }
    
    \IEEEauthorblockA{
        \IEEEauthorrefmark{2}Shanghai Innovation Institute
    }

    \IEEEauthorblockA{
        \IEEEauthorrefmark{3}Li Auto Inc., \{sunbin1, xuhuan5,  haozhihui1, langxianpeng\}@lixiang.com
    }
    
    \href{https://sii-boluomonster.github.io/VecQMDP-website}{\textcolor{orange}{https://sii-boluomonster.github.io/VecQMDP-website}}
    
    % \thanks{\modified{
    % Code and project website are available at:
    % \url{https://github.com/SII-BoluoMonster/VecQMDP} and 
    % \url{https://sii-boluomonster.github.io/VecQMDP-website/}.}}
}

% \and
% \and

\maketitle
\begin{abstract}
    Planning under uncertainty for real-world robotics tasks, such as autonomous driving, requires reasoning in enormous high-dimensional belief spaces, rendering the problem computationally intensive.
    While parallelization offers scalability, existing hybrid CPU-GPU solvers face critical bottlenecks due to host-device synchronization latency and branch divergence on SIMT architectures, limiting their utility for real-time planning and hindering real-robot deployment.
    We present \algname, a CPU-native parallel planner that aligns POMDP search with modern CPUs' SIMD architecture, achieving $227\times$--$1073\times$ speedup over state-of-the-art serial planners. \algname adopts a Data-Oriented Design (DOD), refactoring scattered, pointer-based data structures into contiguous, cache-efficient memory layouts. 
    We further introduce a hierarchical parallelism scheme: distributing sub-trees across independent CPU cores and SIMD lanes, enabling fully vectorized tree expansion and collision checking. Efficiency is maximized with the help of UCB load balancing across trees and a vectorized STR-tree for coarse-level collision checking.
    Evaluated on large-scale autonomous driving benchmarks, \algname achieves state-of-the-art planning performance with millisecond-level latency, establishing CPUs as a high-performance computing platform for large-scale planning under uncertainty.
\end{abstract}

% \begin{IEEEkeywords}
%     POMDP, Parallel Planning, Data-Oriented Design, SIMD Vectorization, Autonomous Driving
% \end{IEEEkeywords}

\IEEEpeerreviewmaketitle

\section{Introduction}

%Paragraph1: 
%Paragraph2: focus of existing work
%Paragraph3: What does our paper proposes

% - What is the problem?
% Show the hand quickly. The problem domain should appear ideally in the first paragraph and definitely no later than the third.

% - Summary of main results
%   - A framework …
%   - An algorithm …
%   - A system …
%   - A theorem …
%   - A set of experiments …

% - Brief discussion of limitations
% Acknowledge limitations and possibly weave future work into the discussion.
Planning under uncertainty for real-world robotics tasks, such as autonomous driving, requires reasoning in enormous high-dimensional belief spaces. In dense urban settings, an autonomous vehicle must navigate interactive traffic flows where surrounding agents exhibit unknown intentions and behaviors. This uncertainty, compounded by perception noise, makes robust planning computationally intensive. Although Partially Observable Markov Decision Processes (POMDPs) provide a principled framework for such problems, the ``curse of dimensionality" and ``curse of history" \cite{somani2013despot} render existing solvers computationally intractable for real-time deployment on edge computing platforms, such as the onboard systems of autonomous vehicles.

While parallelization offers a pathway to scalability, existing solvers face critical bottlenecks on modern hardware. Hybrid CPU-GPU planners, such as HyP-DESPOT \cite{cai2021hyp}, suffer from significant host-device synchronization latency. Alternatively, GPU-native solvers like VOPP \cite{hoerger2025vectorized} are constrained by branch divergence on Single Instruction, Multiple Threads (SIMT) architectures. In autonomous driving, divergence occurs when different scenarios in a belief necessitate distinct logic---for instance, one branch may require yielding while another executes an overtake. This forces the hardware to serialize execution paths, severely degrading throughput and hindering high-frequency, closed-loop control.

% [
% 第三段：
% CPU SIMD is another powerful parallelization method. VAMP has proven it to be highly efficient, i.e., can produce 600x speedup over single-threaded planning.
% However, parralleizing POMDP planning with SIMD is fundamentally more challenging:
%   - the uncertain future renders many different ``scenarios'' or future worlds, each producing a ``scenario tree''. So, need to search multiple trees, not one, whose outcomes together determines the optimal plan.
%   - dynamics is complex. multi-agent interaction renders non-linear trajectories, the computation logic to simulate the dynamics could be complex. Thus it could be hard to vectorize the dynamics function internally.
%   - collision checking is done for a dynamic environment with many agents moving. Spatial trees, typically used in board-phase checking can become different accross scenarios and time steps. It's not possible to pre-compile them.
% Due to the above, vectorizing POMDP planning is hard. It requires a novel formulation that exploits the inherent multi-scenario structure of planning and multi-agent nature of the urban driving problem.
% ]

% 1. 通用图片插入器
% 调用示例: \insertFigure[ht!][0.8\linewidth]{filename}{Caption}{fig:label}{options}
% \NewDocumentCommand{\insertFigure}{O{ht!} O{0.8\linewidth} m m m O{}}{
%     \begin{figure}[#1]
%         \centering
%         \includegraphics[width=#2, #6]{#3}
%         \caption{#4}
%         \label{#5}
%     \end{figure}
% }

\begin{figure}[t]
  \centering
  \includegraphics[width=0.45\textwidth]{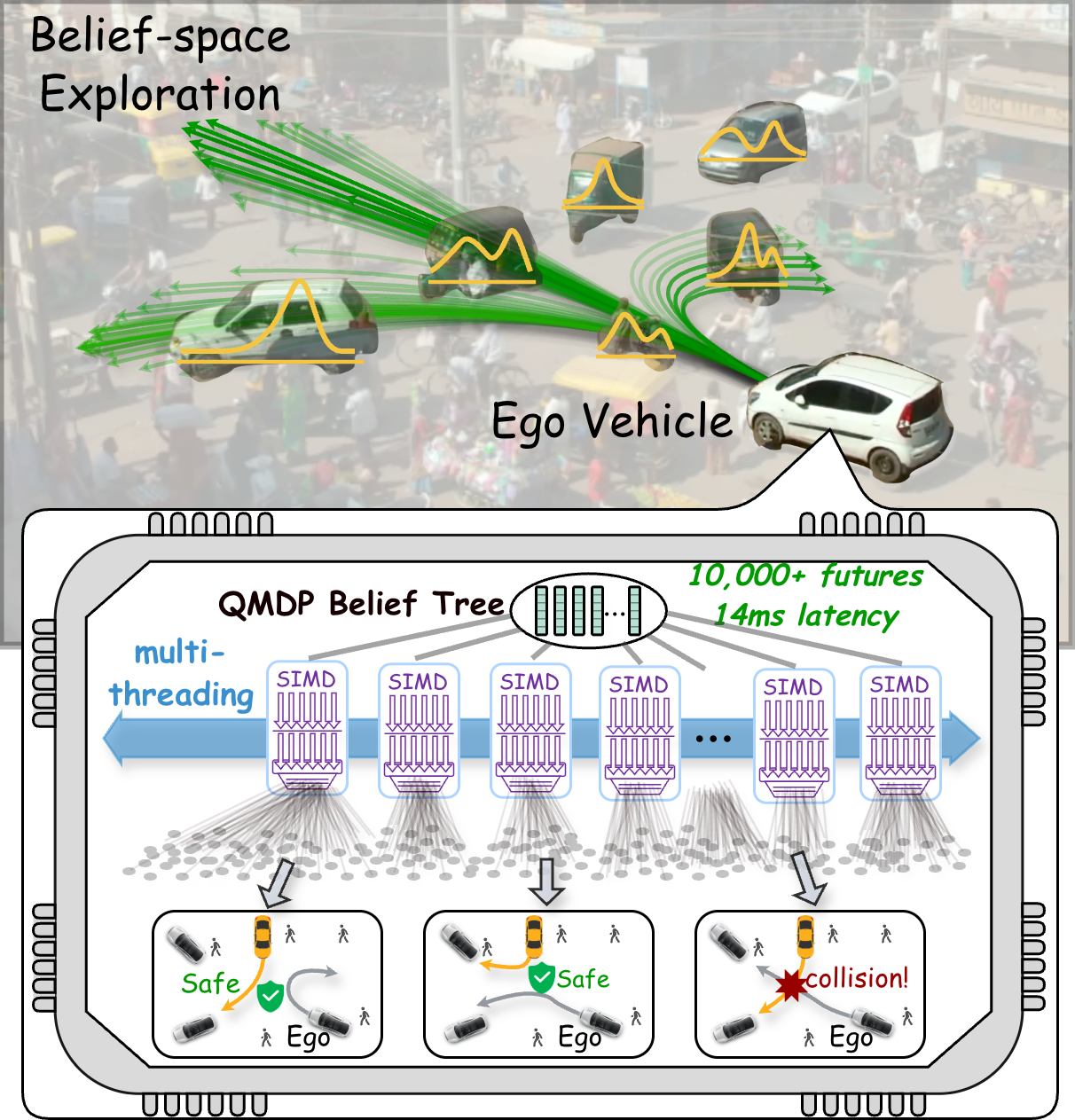}
  \vspace{3.5pt}
    % \caption{\textbf{Real-time belief tree search in complex urban environments.} \algname achieves millisecond-level planning by parallelizing belief tree search across $10,000+$ futures, enabling the \egocar to navigate interactive traffic and respond to high-risk intentions within $14\text{ms}$.}
    \caption{\textbf{Real-time belief-tree search in complex urban environments.} 
    \algname achieves millisecond-level planning by parallelizing belief-tree search over $10{,}000+$ futures, enabling the \egocar to navigate interactive traffic and respond to high-risk interactions within $14\,\mathrm{ms}$.}
  \label{fig:head_fig}
\end{figure}

CPU-based Single Instruction, Multiple Data (SIMD) vectorization is another powerful parallelization paradigm. Recent works like VAMP \cite{thomason2024motions} have proven it to be highly efficient, achieving up to $500\times$ speedups over single-threaded planning in sampling-based motion planning. However, parallelizing POMDP planning with SIMD is fundamentally more challenging. First, the uncertain future renders many different ``scenarios'' or future worlds, each producing a ``scenario tree''. Thus, the solver needs to consider many trees, not one, whose outcomes together determine the optimal plan. Second, the dynamics in large-scale urban environments are complex; multi-agent interaction renders non-linear trajectories, and the computation logic to simulate these dynamics is intricate. Thus, it is difficult to vectorize the dynamics function internally. Third, collision checking is performed for a dynamic environment with many moving agents. Spatial trees, typically used in broad-phase checking, become different across scenarios and time steps, making it impossible to pre-compile them. Due to the above, vectorizing POMDP planning is hard. It requires a novel formulation that exploits the inherent multi-scenario structure of planning and the multi-agent nature of the urban driving problem.

% [
% 第四段：
% to address the above challenges, we propose \algname, a parallel POMDP planner that combines CPU multi-threading and SIMD vectorization. It operates sorely on CPU, thus avoides the communication overhead with GPUs. It has the following core ideas:
%     - it is based on QMDP, an approximate POMDP solver, cite, which decomposes the belief tree into independent sub-trees after the initial action branching, allowing us to distribute the sub-trees across different CPU cores for coarse-grained parallelelization.
%     - to vectorize complex dynamics and collision checking, we introduce two modes of vectorization.
%     - for dynamics, we use ``global vectorization'', i.e., batching the transition dynamics across nodes from different sub-trees, to parallelize the forward simulation of world dynamics in different scenarios.
%     - for collision checking, we use ``local vectorization'', i.e., batching the agents within a single node, to parallelize collision test with the ego-vehicle, including the broad-phase spatial tree traversal and narrow-phase separation test in the corresponding scenario.
%     - to balance workloads across SIMD lanes during global vectorization, we employ a load-balancing UCB mechanism to synchronize the expansion depth of concurrent sub-trees,  to maximize the parallelism.
% ]

To address the above challenges, we propose \algname (\figref{fig:head_fig}), a parallel POMDP planner that combines CPU multi-threading and SIMD vectorization. It operates solely on the CPU, thus avoiding the communication overhead with GPUs. The planner is grounded in the QMDP approximation~\cite{littman1995learning}, which decomposes the belief tree into independent sub-trees after the initial action branching, allowing us to distribute the sub-trees across different CPU cores and SIMD lanes. To vectorize complex dynamics and collision checking, we introduce two modes of vectorization: \textit{global vectorization}, \ie, batching the transition dynamics across nodes from different sub-trees to parallelize the forward simulation of different scenarios; and \textit{local vectorization}, \ie, batching the agents within a single node to parallelize their collision test with the ego-vehicle, including the broad-phase spatial tree traversal and narrow-phase tests. Furthermore, to balance workloads across SIMD lanes during global vectorization, we employ a load-balancing \modified{Upper Confidence Bound (UCB)} to synchronize the expansion-depth range of concurrent sub-trees, thus maximizing parallelism.

% [
%     第五段：
%     experiments：
%     - evaluated on a large-scale autonomous driving benchmark
%     - compared to the state-of-the-art serial POMDP planner for urban driving.
%     - for efficiency, it achieves 227x--1073x speedups in terms of the tree size consturcted in the same planning time.
%     - as to task performance, it acheived state-of-the-art driving performance with millisecond-level planning time, as compared to the serial planner and leading data-driven models, but without requiring any training data.
% ]

We evaluate \algname on the large-scale nuPlan benchmark~\cite{caesar2021nuplan}. In terms of computational efficiency, our framework achieves $227\times$--$1073\times$ speedups in tree construction throughput, defined as the
tree size constructed within unit planning time, compared to a state-of-the-art serial POMDP planner. This massive search capacity enables \algname to deliver state-of-the-art driving performance with millisecond-level planning time, outperforming state-of-the-art learning models without requiring any training data.

% Our core contributions are:
% \begin{itemize}
% \item A hierarchical parallelism scheme that maps the belief forest to hardware across two scales: inter-thread coarse-grained parallelization that distributes independent sub-trees across $M$ CPU cores, and intra-thread fine-grained SIMD vectorization that executes node expansions across $N$ sub-trees simultaneously within each core.
% \item An architectural-aware algorithmic synthesis based on DOD principles. By refactoring search primitives into contiguous, cache-aligned memory layouts, \algname\ enables high-throughput vectorization. This includes a UCB load balancing to maximize SIMD lane utilization and a memory-linearized STRtree for spatial queries via contiguous memory access.
% \item Extensive evaluations on complex urban driving tasks demonstrating that \algname\ achieves state-of-the-art (SOTA) performance, delivering $227\times$--$1073\times$ speedups over existing CPU solvers. The system consistently converges within 16ms, meeting the stringent latency requirements of real-world deployment.
% \end{itemize}

\section{Related Work}
This section reviews efforts to accelerate tree-search-based planning under uncertainty, with an emphasis on how parallelization strategies interact with modern CPU/GPU execution models. We focus on three representative directions: CPU multi-threading for Monte Carlo tree search (MCTS), GPU parallelization for online POMDP planning, and CPU SIMD parallelization inspired by data-oriented layouts.

\subsection{CPU Multi-threading for MCTS-based Planning}
Parallel MCTS commonly uses root, tree, or leaf parallelization~\cite{chaslot2008parallel}. While effective when per-simulation cost is roughly uniform, many robotic tasks violate this assumption. In interactive robotics such as autonomous driving, simulation cost is highly non-uniform, so different threads often traverse different tree paths with unequal workloads, leading to load imbalance and frequent synchronization. Consequently, existing multi-threaded MCTS planners~\cite{chaslot2008parallel, kurzer2020parallelization, cazenave2007parallelization, basu2021parallelizing} typically scale sub-linearly as CPU cores increase in continuous domains.

\subsection{GPU Parallelization for Online POMDP Planning}
GPUs offer high throughput and have motivated both hybrid CPU--GPU pipelines and GPU-resident solvers for online POMDP planning. HyP-DESPOT~\cite{cai2021hyp} accelerates planning by offloading large batches of Monte Carlo simulations to the GPU while retaining higher-level logic-heavy search on the CPU. This partitioning can improve device-side simulation throughput. However, end-to-end planning time may be constrained by host--device communication and by the synchronization required to integrate batched simulation returns across parallel search paths.

More recently, GPU-native planners such as VOPP~\cite{hoerger2025vectorized} avoid host--device communication by keeping data structures and belief tree search on the GPU. However, online POMDP planning in interactive environments often contains data-dependent branching and thus exhibits irregular control flow. This irregularity manifests as \textit{control-flow divergence} on SIMT hardware: when threads within a warp take different branches, the warp executes these paths sequentially via masking, reducing effective utilization and throughput~\cite{han2011reducing}.

\wideFigure[t!][\textwidth]{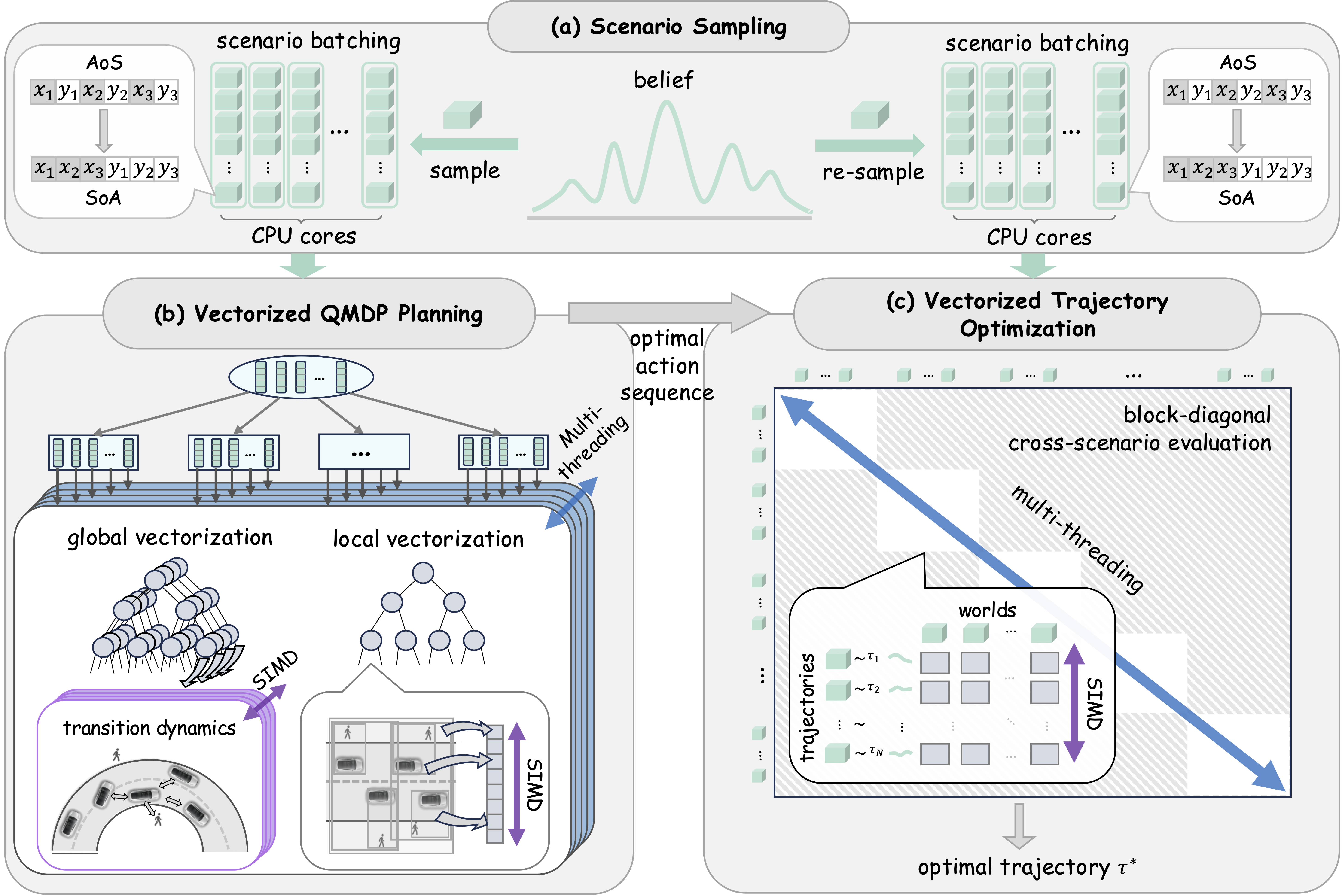}
{\textbf{Overview of \algname.}
(a) Sample the belief into $M\times N$ scenarios in an SoA layout.
(b) Vectorized QMDP search: after the first action, scenario trees run in parallel on $M$ CPU threads; within each thread, SIMD \textit{global vectorization} batches transition dynamics across scenarios and SIMD \textit{local vectorization} accelerates within-node collision checks.
(c) Vectorized trajectory optimization: generate candidates and use block-diagonal cross-scenario evaluation within minibatches to select $\tau^*$.}
{fig:overview}

\subsection{CPU SIMD Parallelization for Tree-search-based Planning}
While modern CPUs provide wide SIMD units, traditional Object-Oriented Programming (OOP)~\cite{cox1984object} layouts are often incompatible with them. Pointer-based structures fragment memory, reduce locality, and break the contiguous access patterns needed for efficient SIMD execution. Recent work such as VAMP~\cite{thomason2024motions} applies Data-Oriented Design (DOD)~\cite{fabian2013data} to sampling-based motion planning in largely static scenes. In particular, Structure of Arrays (SoA)~\cite{strzodka2012abstraction} stores each field contiguously across elements, enabling vectorized loads/stores and higher SIMD utilization.

\modified{Extending SIMD vectorization to online POMDP planning is harder. ROP-RAS3~\cite{liang2024scaling} leverages VAMP-style SIMD acceleration to sample diverse macro-actions, improving scalability toward extremely long horizons. In contrast, \algname targets real-time autonomous driving with a comprehensive CPU-native DOD framework that vectorizes the core planning cycle, including belief-tree expansion, dynamics simulation, and collision checking. This setting is especially challenging because the planner must evaluate many scenario trees rather than a single rollout. Urban driving also involves complex multi-agent interactions, causing data-dependent branching in the transition model, while dynamic collision checking requires spatial trees that vary across scenarios and time steps and therefore cannot be pre-compiled.

\algname tightly couples multi-threading with SIMD under the QMDP approximation~\cite{littman1995learning}. It applies \textit{global vectorization} to batch transition dynamics across scenarios and \textit{local vectorization} to accelerate within-node multi-agent collision checks, achieving up to a $1073\times$ speedup over state-of-the-art serial solvers.}

\section{Overview}

\algname (\figref{fig:overview}) scales up a state-of-the-art POMDP planner \hidrive~\cite{jin2025hi} for autonomous driving by leveraging SIMD parallelism, demonstrating how belief tree search and belief-space trajectory optimization can be extensively vectorized for robotics tasks in complex dynamic environments. We build our parallel planner upon the QMDP approximation, which decomposes the belief tree into a set of independent sub-trees after the initial action branching, where each sub-tree corresponds to a sampled scenario of the future world. This decomposition allows us to distribute minibatches of scenario trees across different CPU cores for coarse-grained parallelization.

We then vectorize the QMDP belief tree search (\figref{fig:overview}b) using two distinct modes. \textit{Global vectorization} batches search components across different scenario trees and assigns them to different SIMD lanes for parallel execution. This mode is applied to components with complex logic that are hard to vectorize internally, such as the transition dynamics of multi-agent interactions. \textit{Local vectorization} vectorizes the internal computation of a search component. This mode is applied to components with clear factorization structure, such as collision checking between the \egocar and independent exogenous agents. To support these vectorization modes, we refactor our data structures and algorithmic logic according to  Data-Oriented Design (DOD) principles. We transform pointer-based trees into \textit{vectorized trees} to enable SIMD-based traversal and convert Arrays of Structures (AoS) into Structures of Arrays (SoA) to ensure contiguous memory access.

Following the belief tree search, \algname employs belief-space trajectory optimization (\figref{fig:overview}c) to refine driving trajectories based on the optimal action sequence. The core computation involves cross-scenario evaluation, where the \egocar trajectory generated in each scenario is evaluated against the external world in other scenarios. We organize this evaluation as a block-diagonal sparse matrix, where each block concerns a minibatch of scenarios processed in a vectorized SIMD manner. Here, the vectorization is predominantly global, as the transition dynamics and collision checking of different scenarios are collated and vectorized.

The remainder of this paper is organized as follows: \secref{sec:POMDP_and_QMDP} formulates the POMDP model and the QMDP approximation. \secref{sec:vec_bts} details our vectorized QMDP planner. \secref{sec:vec_bos} introduces our vectorized belief-space trajectory optimization. \secref{sec:experiments} presents experimental results the large-scale nuPlan benchmark~\cite{caesar2021nuplan} and discussions.

\section{Mathematical Foundation for Parallelization} \label{sec:POMDP_and_QMDP}
This section formalizes the POMDP model for autonomous driving and introduces the QMDP approximation~\cite{littman1995learning}, which provides the mathematical foundation for parallelization.

\subsection{POMDP Model}
We model driving as a POMDP $\langle \mathcal{S},\mathcal{A},\mathcal{O},\mathcal{T},\mathcal{Z},\mathcal{R}, b_0, \gamma \rangle$.
A state $s \in \mathcal{S}$ includes the physical states of all agents (ego and exogenous), such as positions, speeds, headings, geometry, \etc \
Uncertainty arises from the behavioral intentions of exogenous agents, represented as predicted future trajectories $\boldsymbol{\xi}=\{\xi^i\}_{i=1}^{n_\mathrm{exo}}$ on the planning horizon.
We maintain a belief $b(\boldsymbol{\xi}\mid s)$ from a multimodal trajectory predictor (\eg, QCNet~\cite{zhou2023query}), capturing multiple plausible future worlds conditioned on the current states.

The \egocar selects a macro-action $a \in \mathcal{A}$ from a discrete set of reference paths with lateral nudges.
Concretely, we use $3$ candidate reference paths (\eg, lane-following, adjacent-lane change) and apply three lateral nudges of -1 m, 0 m, and +1 m to each path, yielding $|\mathcal{A}|=9$ macro-actions.
Each macro-action spans $\Delta t=2$s with a finite planning horizon $T=8$s.
Conditioned on sampled multi-agent trajectory realizations $\boldsymbol{\xi}$, the transition advances the joint scene state by multi-agent simulation:
$s' \sim \mathcal{T}(s' \mid s, a; \boldsymbol{\xi})$.
Observations are noisy sensor readings modeled by $o \sim \mathcal{Z}(o \mid s')$, and the reward $\mathcal{R}(s,a)$ trades off safety, efficiency, and comfort with discount $\gamma \in [0,1]$.

\subsection{QMDP Approximation} \label{sec:QMDP_solver}
% To support real-time planning, \algname\ adopts the QMDP approximation~\cite{littman1995learning}, which assumes uncertainty is resolved after the \emph{first} step. 
% We sample $K$ scenarios $\phi=(s,\boldsymbol{\xi})$ from the current belief; each $\phi$ fixes a \textit{deterministic} rollout for all exogenous agents. 
% Conditioned on $\phi$, the problem corresponds to solving a fully observable MDP, and we compute the value function $V_\phi(\cdot)$ by lookahead search over the corresponding scenario tree.

To support real-time planning, \algname\ adopts the QMDP approximation~\cite{littman1995learning}, which assumes uncertainty is resolved after the \emph{first} step. 
\modified{We sample $K$ scenarios $\phi=(s,\boldsymbol{\xi})$ from the current belief, where each scenario specifies a sampled initial state and determinizes the future evolution of exogenous agents, transitions, and observations.}
Conditioned on $\phi$, the problem reduces to a fully observable MDP, and we compute the value function $V_\phi(\cdot)$ by lookahead search over the corresponding scenario tree.

Under QMDP, the root action value under belief $b$ is approximated by averaging the post-action MDP values across $K$ fixed scenarios, where $s'_k = \mathcal{T}(s, a; \boldsymbol{\xi}_k)$ is deterministic given $(s,a,\boldsymbol{\xi}_k)$:
\begin{align}
Q_{\mathrm{QMDP}}(b, a) &\approx \frac{1}{K}\sum_{k=1}^{K}
\left[
R(s, a) + \gamma V_{\phi_k}(s'_k)
\right], \label{eq:qmdp_approximation} \\
a^* &= \arg\max_{a \in \mathcal{A}} Q_{\mathrm{QMDP}}(b,a). \label{eq:optimal_action}
\end{align}

This reveals the key parallel structure: after the first action branching, the $K$ scenario trees become independent and can be solved in parallel. The resulting values are then aggregated to determine the optimal root action $a^*$. We recover the optimal action sequence $\pi^*$ by following the highest-value action branches within the scenario trees. \algname leverages this structure for hierarchical parallelization in \secref{sec:vec_bts} and refines $\pi^*$ using belief-space trajectory optimization in \secref{sec:vec_bos}.

\section{Vectorized QMDP Belief Tree Search} \label{sec:vec_bts}
This section presents a CPU-native realization of hierarchical parallelism for QMDP belief tree search. Building on the QMDP decomposition in \eqref{eq:qmdp_approximation}, we first describe the belief tree search in \algoref{alg:qmdp_tree}, and then show how its \textsc{Expansion}/\textsc{Rollout} and reward evaluation are implemented with SIMD vectorization.

\begin{algorithm}[!t]
\caption{Vectorized QMDP Belief Tree Search}
\label{alg:qmdp_tree}
\begin{algorithmic}[1]
\REQUIRE Initial belief $b_0$, \modified{number of CPU threads $M$, scenarios per thread $N$}
\ENSURE Optimal action sequence $\pi^*$
\STATE $\Phi = \{\phi_1, ..., \phi_K\} \gets \text{SampleScenarios}(b_0)$, \modified{where $K=MN$}
\STATE Expand root over all $a \in \mathcal{A}$
\STATE Initialize scenario trees $\mathcal{F}=\{\mathcal{T}^{(1)}_{\mathrm{sc}},\dots,\mathcal{T}^{(K)}_{\mathrm{sc}}\}$ conditioned on $\Phi$ after first action branching
\FORALL{thread $m \in \{1, \dots, M\}$ \textbf{in parallel}}
    \STATE $\mathcal{F}_m \gets \{\mathcal{T}^{(m-1)N + 1}_{\mathrm{sc}}, \dots, \mathcal{T}^{mN}_{\mathrm{sc}}\}$
    \WHILE{within planning time budget}
        \STATE $\text{Traverse}(\mathcal{F}_m)$
        \STATE $\mathcal{F}_m \gets \text{VectorizedExpansion}(\mathcal{F}_m)$
        \STATE $\text{VectorizedRollout}(\mathcal{F}_m)$
        \STATE $\text{BackUp}(\mathcal{F}_m)$
        \IF{$\text{Converged}(Q\text{-values})$}
            \STATE \textbf{break}
        \ENDIF
    \ENDWHILE
\ENDFOR
\RETURN $\pi^* \gets \text{ExtractActionSequence}(\mathcal{F})$
\end{algorithmic}
\end{algorithm}

\subsection{QMDP Planning for Urban Driving}\label{sec:qmdp_planning_urban}
\algoref{alg:qmdp_tree} performs finite-horizon belief-tree search over the discrete macro-action set $\mathcal{A}$ under $K$ sampled scenarios. It samples $\Phi=\{\phi_k\}_{k=1}^{K}$ from the initial belief $b_0$, expands the root over all $a\in\mathcal{A}$, and decomposes the belief tree into $K$ independent scenario trees $\mathcal{F}=\{\mathcal{T}^{(1)}_{\mathrm{sc}},\dots,\mathcal{T}^{(K)}_{\mathrm{sc}}\}$, each conditioned on one $\phi_k$. 
The search then repeats \textsc{Traverse} to select a node, applies \textsc{Expansion} to execute one macro-action, runs \textsc{Rollout} to simulate to horizon $H$, and performs \textsc{BackUp} to update $Q(v,a)$, $N(v,a)$, and the \textit{expansion-depth range} required by the load-balancing UCB in \secref{sec:lb_UCB}. Planning stops when the root $Q$-values converge or the time budget is reached, and extracts $\pi^*$ from the aggregated root $Q$-values under QMDP.

\textsc{Expansion}/\textsc{Rollout} simulates the joint evolution conditioned on the sampled $\boldsymbol{\xi}$. Exogenous agents deterministically follow the sampled trajectories $\boldsymbol{\xi}$, while the \egocar executes a macro-action by following its reference path with closed-loop interaction: longitudinal behavior follows \modified{Intelligent Driver Model (IDM)} and lateral control uses a Stanley controller~\cite{treiber2000congested,thrun2006stanley}, with lane changes gated by a \modified{Minimizing Overall Braking Induced by Lane changes (MOBIL)} feasibility check~\cite{kesting2007general}. This simulation is further accelerated in \secref{sec:global_vectorization}.

Rewards are safety-dominated and evaluated against the same $\boldsymbol{\xi}$. For broad-phase filtering, we build a Frenet-based Sort-Tile-Recursive (STR)-tree~\cite{leutenegger1997str} from predicted agent geometries at each time step and reuse it during search since $\boldsymbol{\xi}$ is fixed at the root. Broad-phase queries test the ego Frenet-frame  Axis-Aligned Bounding Box (AABB) \cite{cai2013collision} against STR-tree node AABBs to prune candidates; leaf hits return a compact set of agent indices. We then gather the corresponding Cartesian states and apply a narrow-phase \modified{Separating Axis Theorem (SAT)} test~\cite{huynh2009separating} to compute collisions, penalties, and terminal conditions. Frenet-frame AABBs align with road heading, reducing bounding-box inflation on curved roads and thus false positives.
This reward pipeline is SIMD-vectorized in \secref{sec:local_vectorization}.

\subsection{Data-Oriented Representations for Vectorized Search} \label{sec:dod_vectorized_search}
To enable SIMD processing, we refactor two pointer-based structures into SIMD-friendly layouts: the scenario search trees and the spatial STR-tree. Both have \emph{fixed} maximum sizes, so we pre-allocate them as array-based balanced hierarchies with padding for absent nodes. For scenario trees, the capacity is fixed by planning depth $H$ and branching $|\mathcal{A}|$, yielding $N_{\mathrm{sc}}$ (e.g., $N_{\mathrm{sc}}=\sum_{d=0}^{H}|\mathcal{A}|^{d}$). For STR-trees, the maximum leaf count is bounded by the number of exogenous agents $n_{\mathrm{exo}}$ and a fixed branching factor, yielding $N_{\mathrm{str}}$. Each CPU thread owns a disjoint subset of scenarios and STR-trees, so all structures are built \textit{intra-thread} without locks.

For each scenario tree, we flatten nodes into contiguous indices $v\in\{0,1,\dots,N_{\mathrm{sc}}-1\}$ and encode topology implicitly with fixed branching. For an action index $i \in \{1,\dots,|\mathcal{A}|\}$, the $i$-th child of node $v$ is addressed as $|\mathcal{A}|\cdot v + i$. We use a Structure-of-Arrays (SoA) layout for per-node statistics and flags, including $Q$-values, visit counts, immediate reward, and terminal/expanded markers. We also cache the \emph{ego state} at each node, i.e., the kinematic state reached after executing the macro-action prefix along the root-to-$v$ path. Concretely, ego states are stored as SoA vectors over nodes (\eg, $\{x^{v}\}_{v=0}^{N_{\mathrm{sc}}-1}$)
Exogenous trajectories are scenario-level inputs fixed at the root under QMDP, so we store them once per scenario in a time-major layout, using SoA over agents at each time step (\eg, for each $t$, $\{x^{i}_{t}\}_{i=1}^{n_{\mathrm{exo}}}$).

The STR-tree is stored as an array-based balanced hierarchy in contiguous buffers, again pre-allocated with padding. We store AABB fields in SoA form (e.g., $\min_s,\max_s,\min_d,\max_d$ in the Frenet frame), so SIMD lanes can load multiple child boxes and test them against the ego AABB in parallel during broad-phase filtering.

\subsection{Global Vectorization Across Scenario Trees}\label{sec:global_vectorization}
Global vectorization accelerates \textsc{Expansion}/\textsc{Rollout} \emph{across scenarios} by batching $N$ expanded nodes drawn from $N$ independent scenario trees within each CPU thread. Given the selected expanded node indices $\{v_i\}_{i=1}^{N}$, we \textit{SIMD-gather} the corresponding ego states from SoA buffers into SIMD vector registers (e.g., $\{x^{v_i}\}_{i=1}^{N}, \{y^{v_i}\}_{i=1}^{N}$), together with per-node metadata (depth, terminal flags) and the macro-action IDs. Exogenous trajectories are scenario-level inputs fixed at the root, so each SIMD lane reads aligned per-time-step agent states from a time-major layout, with an SoA organization over agents at each time step.

In \textsc{VectorizedExpansion}, we execute the selected macro-action for all $N$ SIMD lanes using a single step-synchronous kernel. At each simulation step, the kernel advances all SIMD lanes in lockstep by loading the required per-SIMD-lane ego states, evaluating per-SIMD-lane control updates with masked execution, and writes back the updated ego states to the corresponding node buffers. The control flow remains largely uniform due to the macro-action design, with divergence occurring only when a macro-action triggers a lane change. In this case, only the affected SIMD lanes perform a MOBIL path-change feasibility check, while others are masked.

\textsc{VectorizedRollout} reuses the same step-synchronous kernel and differs only in termination. SIMD lanes are progressively masked once they reach horizon $H$ or trigger a terminal event, and the rollout continues until all SIMD lanes are masked. Since expanded nodes can start at different depths (\emph{depth divergence}), SIMD lanes may terminate at different times while the SIMD group runs to the last active SIMD lane, motivating load-balanced selection in \secref{sec:lb_UCB}.

\subsection{Local Vectorization Within a Scenario Node}\label{sec:local_vectorization}
Local vectorization accelerates reward evaluation by SIMD-parallelizing collision checking \emph{within} a scenario node, as illustrated in \figref{fig:frenet_collision_check}. We batch multiple exogenous agents into one SIMD group and process them in lockstep.

In broad-phase filtering (\figref{fig:frenet_collision_check}a), STR-tree traversal is SIMD-accelerated at each expansion/rollout step. For an intersected internal node, we \textit{batch-load} child AABBs from SoA buffers into SIMD vector registers and perform vectorized AABB intersection tests against the \egocar's AABB. We descend only into intersecting children; leaf hits return candidate agent indices, which we then \textit{SIMD-gather} into SIMD vector registers as compact blocks of exogenous states (e.g., position, heading, geometry) for refinement.

In narrow-phase refinement (\figref{fig:frenet_collision_check}b), we apply a SIMD Separating Axis Theorem (SAT) kernel, where each SIMD lane evaluates one ego--agent pair. We use masking with \textit{early exit}: a SIMD lane stops once a separating axis is found, and the kernel terminates when all SIMD lanes exit or all SAT axes are tested. The kernel outputs collision flags and contact outcomes for rewards and termination.

\subsection{Selection with Load-Balancing UCB}\label{sec:lb_UCB}
To mitigate the load imbalance noted in \secref{sec:global_vectorization}, we modify \textsc{Traverse} (Alg.~\ref{alg:qmdp_tree}, line~7) to reduce depth divergence across scenario trees in different SIMD lanes.

Each node $s$ maintains an expansion-depth range $\mathcal{D}(s)=[d_{\min}(s),d_{\max}(s)]$, where $d_{\min}$ and $d_{\max}$ represent the minimum and maximum depths of all unexpanded expanded nodes in the subtree rooted at $s$. A new node initializes $\mathcal{D}(s)$ based on its own depth. During \textsc{BackUp} (Alg.~\ref{alg:qmdp_tree}, line~10), $\mathcal{D}(s)$ is updated by taking the union of the current node's depth range and the depth ranges of its child nodes, terminating early if no change occurs.

Within each SIMD batch, selection proceeds in two stages. We first run standard UCB \cite{auer2002finite} independently in each scenario tree to obtain a tentative expanded node and record its depth; a majority vote over these depths defines the reference depth $d_{\mathrm{ref}}$. We then re-run selection using a load-balancing UCB. For an action $a$ at node $s$ with child $s'$, we score:
\begin{equation*}
\mathcal{U}(s, a) = \text{UCB}(s, a) - \lambda \cdot \left| \text{clamp}(d_{\mathrm{ref}}, \mathcal{D}(s^\prime)) - d_{\mathrm{ref}} \right|,
\end{equation*}
where larger $\lambda$ increasingly biases selection toward children whose expansion-depth range contains (or is closest to) $d_{\mathrm{ref}}$, aligning selected depths across scenario trees for better load balancing.

\begin{table*}[!t]
    \centering
    \caption{Driving Performance Comparison on nuPlan.}
    \renewcommand{\arraystretch}{1.5}
    \setlength{\extrarowheight}{-1pt}
    \setlength{\tabcolsep}{0.65em} % 增加列间距
    \begin{tabular}{llccccccc}
        \toprule
        \multirow{2}{*}{\textbf{Type}} & \multirow{2}{*}{\textbf{Planner}} & \multicolumn{2}{c}{\textbf{Val14}} & \multicolumn{2}{c}{\textbf{Test14-random}} & \multicolumn{2}{c}{\textbf{Test14-hard}} & \multirow{2}{2cm}{\centering \textbf{Inference / Planning Time ($\mathrm{ms}$)} \blackdown} \\
        \cmidrule(lr){3-4} \cmidrule(lr){5-6} \cmidrule(lr){7-8}
        & & \textbf{R} & \textbf{NR} & \textbf{R} & \textbf{NR} & \textbf{R} & \textbf{NR} & \\\midrule
        \textcolor{gray}{\textit{Expert}} & \textcolor{gray}{Log-replay} & \textcolor{gray}{80.32} & \textcolor{gray}{93.53} & \textcolor{gray}{75.86} & \textcolor{gray}{94.03} & \textcolor{gray}{68.80} & \textcolor{gray}{85.96} & \textcolor{gray}{-} \\
        \midrule
        \multirow{2}{*}{Learning-based} 
        & PLUTO & 78.11 & 88.89 & 78.62 & 89.90  & 59.74 & 70.03 & - \\
        & Diffusion Planner & 82.86\ci{0.14} & 89.67\ci{0.05} & 84.60\ci{0.16} & 89.71\ci{0.32} & 68.77\ci{0.15} & 75.30\ci{0.31} & 80 \\
        \midrule
        \multirow{3}{*}{Hybrid} 
        & PDM-Hybrid & 92.14\ci{0.00} & 92.77\ci{0.00} & 91.79\ci{0.00} & 94.56\ci{0.00} & 76.12\ci{0.00} & 66.09\ci{0.00} & 171 \\
        & PLUTO w/ refine. & 76.88 & 92.88 & 90.29 & 92.23 & 76.88 & 80.08 & - \\
        & Diff. Planner w/ refine. & 92.90 & \underline{94.26} & 91.75 & \underline{94.80} & 82.00 & 78.87 & $>$80 \\
        \midrule
        \multirow{4}{*}{Model-based}
        & \hidrive & 93.01\ci{0.12} & 93.22\ci{0.08} & 91.95\ci{0.13} & 93.46\ci{0.31} & \underline{83.07}\ci{0.18} & 81.22\ci{0.31} & 503 \\
        & Hyp-DESPOT & 93.07\ci{0.06} & 93.26\ci{0.13} & 91.83\ci{0.15} & 93.69\ci{0.20} & 82.83\ci{0.29} & 81.82\ci{0.28} & 259 \\
        & \algname (match, Ours) & \underline{93.15}\ci{0.11} & 94.16\ci{0.03}& \underline{92.51}\ci{0.00} & \textbf{95.21}\ci{0.00} & \textbf{84.23}\ci{0.35} & \underline{82.30}\ci{0.48}  & 9 \\
        & \algname (best, Ours) & \textbf{93.22}\ci{0.06} & \textbf{94.36}\ci{0.02} & \textbf{93.04}\ci{0.05} & \textbf{95.21}\ci{0.00} & \textbf{84.23}\ci{0.35} & \textbf{82.84}\ci{0.11}  & 14 \\
        \bottomrule
    \end{tabular}
    \label{table::comparision_baselines}
    \hspace*{-0.05\linewidth}
    % \parbox{0.9\linewidth}{
    %     \vspace{0.15cm}
    %     \small * \emph{\textbf{Bold} indicates best; \underline{underscored} indicates second-best.}
    % }
    \parbox{0.9\linewidth}{
        \vspace{0.15cm}
        \footnotesize
        \small * \emph{\textbf{Bold} indicates best; \underline{underscored} indicates second-best.}\\
        \emph{Entries without confidence intervals are reported as published in the corresponding original papers.}
    }
    \vspace{-0.3cm}
\end{table*}
\section{Vectorized Belief-space Trajectory Optimization} \label{sec:vec_bos}
Leveraging the optimal action sequence from the belief tree search, \algname refines a continuous trajectory that is robust under scenario uncertainty.

\subsection{Belief-space Trajectory Optimization with Importance Sampling}
To focus optimization on safety-critical interactions, we re-sample $K$ scenarios using importance sampling \cite{luo2019importance}. Risk-relevant agents are identified from map topology and predicted intersections between their future trajectories and the \egocar's reference path. A proposal distribution $q$ shifts probability mass of these critical agents toward hazardous intentions, while the nominal belief $b$ is retained for non-critical agents.

For each sampled scenario, a candidate ego trajectory is generated by forward simulation under the action sequence, following the transition dynamics described in \secref{sec:qmdp_planning_urban}. Robust evaluation requires \emph{cross-scenario evaluation}: for each trajectory–scenario pair $(\tau_k,\phi_i)$, we simulate the world forward conditioned on executing $\tau_k$ and evaluate the resulting state sequence using the reward model from \secref{sec:qmdp_planning_urban}, yielding $V(\tau_k \mid \phi_i)$. The expected value of a trajectory is estimated via self-normalized importance sampling \cite{hesterberg1988advances}:

\begin{equation*}
\begin{aligned} 
E[V(\tau_k)] &= \frac{1}{\sum_{i=1}^{K} w_i}\sum_{i=1}^{K} w_i\, V(\tau_k \mid \phi_i), 
\\
\tau^* &= \arg\max_{\tau_k} E[V(\tau_k)],
\end{aligned} 
\end{equation*}
where $w_i=\prod_{j} \frac{b(\xi_{i,j})}{q(\xi_{i,j})}$ is the product of importance weights over all agents $j$ in scenario $\phi_i$.

\insertFigure[!t][\columnwidth]{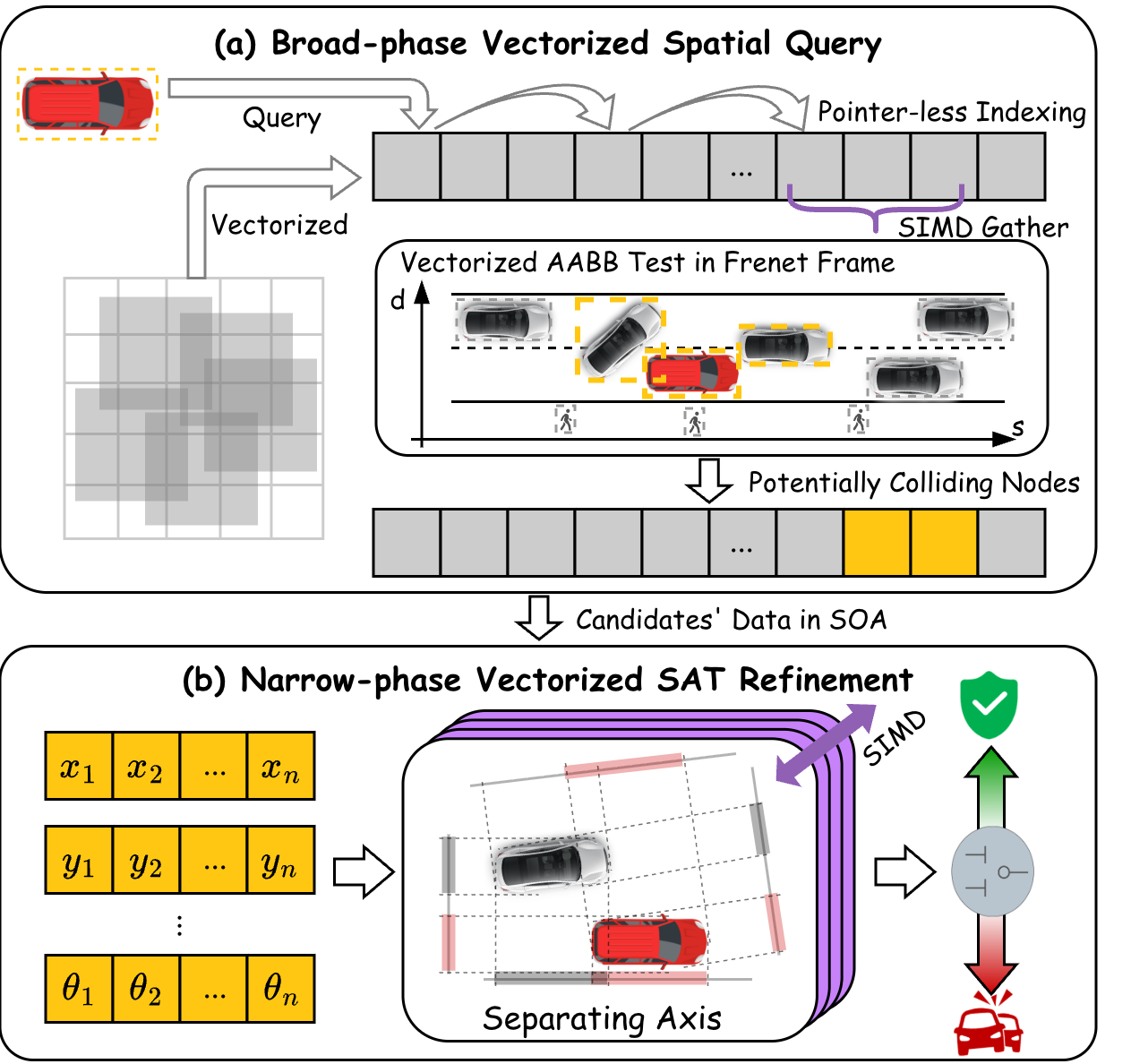}
{\textbf{Two-stage SIMD collision checking.}
(a) Broad phase: SIMD AABB tests in the Frenet frame traverse a pointer-less STR-tree to prune candidates.
(b) Narrow phase: SIMD SAT checks evaluate ego--agent pairs to compute collisions.}
{fig:frenet_collision_check}

\subsection{Parallel Block-Diagonal Cross-Scenario Evaluation}
A full cross-scenario evaluation scales as $O(K^2)$ and is intractable in real time. To mitigate this, we partition the $K$ scenarios across $M$ CPU threads, where each thread handles a minibatch $\Phi_m$ of $N$ scenarios, restricting cross-evaluation within the minibatch and reducing per-thread complexity to $O(N^2)$.

Within each thread, computation is organized around \textit{global vectorization} across trajectories. Ego trajectories for all $N$ scenarios in $\Phi_m$ are generated in parallel via SIMD-batched forward simulation, as described in \secref{sec:global_vectorization}. During evaluation, trajectory points across this SIMD batch are processed together: each trajectory is evaluated across all $N$ scenarios in $\Phi_m$, assessing their respective future states. Reward computation is dominated by collision checking. For each time step, the batched ego geometries form a query region that is intersected with the scenario’s STR-tree to obtain candidate agents. Vectorized narrow-phase checks are then performed, with each check comparing multiple ego trajectories to each candidate exogenous agent. This structure enables efficient parallelization across both ego trajectories and exogenous agents during collision checking, making belief-space optimization tractable on CPU platforms.

\section{Experiments} \label{sec:experiments}
We evaluate \algname on the large-scale nuPlan benchmark~\cite{caesar2021nuplan} to validate its performance in complex, interactive traffic scenarios under strict real-time constraints. Our results demonstrate that \algname achieves competitive driving performance, with planning times as low as 9ms and peak scores within a 14ms planning time. In terms of computational efficiency, \algname delivers a $227\times$–$1073\times$ speedup in belief tree construction throughput compared to the state-of-the-art serial POMDP planner \hidrive, with the gains scaling proportionally to scene complexity. While the evaluation focuses on autonomous driving, the core principles of CPU-parallelized belief tree search—leveraging multi-threading and SIMD to handle high-dimensional belief spaces and complex multi-agent dynamics—are applicable to other robotics domains requiring similar reasoning capabilities.

\subsection{Experimental Setting}
We evaluate \algname on the nuPlan benchmark~\cite{caesar2021nuplan} using three subsets of increasing difficulty: \textit{Val14}~\cite{Dauner2023CORL} (1,118 regular scenes), \textit{Test14-random}~\cite{cheng2024rethinking} (268 random scenes), and \textit{Test14-hard}~\cite{cheng2024rethinking} (272 complex scenes). 
Each scene involves a 15-second closed-loop simulation, evaluated in both Non-Reactive (NR) (static log-replay) and Reactive (R) (interactive agents) modes. 
\modified{All test scenes are directly taken from nuPlan without modification; dense scenes with large numbers of agents naturally arise from urban traffic involving vehicles, pedestrians, and cyclists.}

\modified{
Baselines include contrastive imitation learning (PLUTO~\cite{cheng_pluto_2024}), diffusion-based planning (Diffusion Planner~\cite{zheng2025diffusionbased}), hybrid optimization (PDM-Hybrid~\cite{Dauner2023CORL}), the state-of-the-art serial POMDP planner \hidrive~\cite{jin2025hi}, and HyP-DESPOT~\cite{cai2021hyp}, a CPU multi-threaded parallel variant of \hidrive. All methods are evaluated under identical hardware settings. Reported runtimes measure planning only and exclude prediction and perception, since nuPlan provides processed geometric and kinematic states. For a fair comparison among POMDP planners, \hidrive, HyP-DESPOT, and \algname use the same $K=64$ sampled scenarios, and both HyP-DESPOT and \algname are run with $8$ CPU threads.
}

\modified{
Experiments are conducted on an Intel Xeon Gold 6530 CPU, where \algname uses AVX2 with SIMD width $N=8$. The implementation also supports AVX-512 on x86 and NEON on ARM platforms, making the proposed Data-Oriented Design and SIMD vectorization transferable to embedded automotive SoCs such as NVIDIA Orin. Prior SIMD results on Jetson AGX Orin suggest an approximate $2.2\times$ slowdown relative to our AVX2 setting~\cite{wei2025t}, indicating strong potential for lightweight onboard deployment.
}

\subsection{Evaluation Metrics}
We evaluate planning quality and computational performance using the following metrics:

% \textbf{Driving Score}: We use the standard nuPlan scoring function to measure closed-loop driving performance, aggregating safety and rule-compliance indicators such as collision avoidance, goal progress, and adherence to road boundaries.
\modified{
\textbf{Driving Score and Safety Metrics}: We use the standard nuPlan scoring function to measure closed-loop driving performance, which aggregates safety, progress, and rule-compliance indicators. In addition to the overall Driving Score (DS), we report key safety-related nuPlan metric scores: Safety Score (SS), corresponding to the not-at-fault collision score; TTC Score (TTCS), which evaluates whether the minimum time-to-collision satisfies the prescribed safety bound; and Drivable Area Compliance Score (DACS), which evaluates whether the ego vehicle remains within the mapped drivable area.
}

\modified{
\textbf{Tree Construction Throughput}: To quantify computational capacity, we define \textit{tree construction throughput} as the number of constructed tree edges per unit planning time (edges/ms). Each expanded edge corresponds to one \textit{future}, \ie, one simulated transition branch under a sampled scenario; thus, expanding a node at depth $d$ with maximum horizon $H$ contributes $H-d$ futures. For a fair comparison, both \hidrive and HyP-DESPOT use the same QCNet-predicted trajectories as \algname, and their edge counts are weighted by the number of scenarios visiting each edge. Since the per-edge processing time in \algname is stable across the tree (mean $0.038$ ms, std. $0.013$ ms), each edge can be treated as an atomic unit, making edges/ms a robust measure of real-time search capacity.
}

\begin{table}[t]
\centering
\footnotesize % 适当缩小字号以适应单栏
\caption{Safety Performance Comparison on nuPlan.}
\label{table::safety_metrics}
\renewcommand{\arraystretch}{1.0}
\setlength{\tabcolsep}{2pt} % 减小列间距

% 使用 tabularx，总宽度设定为 \columnwidth
% 定义一个新的列类型 Y，使其内容居中且等宽分配空间
\newcolumntype{Y}{>{\centering\arraybackslash}X}

\begin{tabularx}{\columnwidth}{@{} l YYY YYY @{}}
\toprule
& \multicolumn{3}{c}{\textbf{Val14} (NR)} & \multicolumn{3}{c}{\textbf{Test14-Hard} (NR)} \\
\cmidrule(lr){2-4} \cmidrule(lr){5-7}
\textbf{Method} & SS$\uparrow$ & TTCS$\uparrow$ & DACS$\uparrow$ & SS$\uparrow$ & TTCS$\uparrow$ & DACS$\uparrow$ \\
\midrule
HyP-DESPOT & 98.74 & 90.79 & 100.00 & 96.32 & 83.46 & 98.90 \\
\algname    & 99.36 & 93.91 & 100.00 & 96.76 & 83.79 & 98.16 \\
\bottomrule
\end{tabularx}
\end{table}

\subsection{Comparison on Driving Performance} \label{sec:performance}
As shown in Table \ref{table::comparision_baselines}, \algname outperforms learning-based, model-based, and hybrid baselines across all nuPlan scenes. We report two planning-time budgets to assess the efficiency--performance trade-off.

\modified{
\algname (match) achieves performance parity with state-of-the-art methods within only 9\,$\mathrm{ms}$, while \algname (best) reaches peak driving performance within 14\,$\mathrm{ms}$. At the 14\,$\mathrm{ms}$ budget, runtime is dominated by expansion, rollout, and trajectory optimization (3.23\,$\mathrm{ms}$, 4.05\,$\mathrm{ms}$, 4.68\,$\mathrm{ms}$), while other components remain lightweight, including scenario sampling (0.86\,$\mathrm{ms}$), STR-tree construction (0.99\,$\mathrm{ms}$), selection (0.28\,$\mathrm{ms}$), and backup (0.17\,$\mathrm{ms}$). Compared with HyP-DESPOT, \algname achieves 1683.49 versus 10.88 edges/ms, corresponding to a $154.73\times$ average tree-construction-throughput speedup. The safety results in~\tabref{table::safety_metrics} further show that \algname improves Safety Score and TTC Score over HyP-DESPOT on both Val14 and Test14-hard, suggesting that the QMDP approximation preserves safety relative to full POMDP planning. Including QCNet prediction latency, \algname still achieves an end-to-end latency of 65\,$\mathrm{ms}$, lower than Diffusion Planner's 80\,$\mathrm{ms}$, while obtaining better driving performance.

Performance gains are most pronounced on Test14-hard. In these challenging scenes, the high tree-construction throughput enables \algname to evaluate many futures within 14\,$\mathrm{ms}$, allowing it to detect and negotiate low-probability, high-risk interactions, such as aggressive cut-ins, that are difficult for lower-throughput planners to capture.
}

\subsection{Computational Throughput Comparison} \label{sec:throughput}
\algname's computational efficiency is evaluated by analyzing tree construction throughput across varying traffic densities. \figref{fig::throughput_comparison} shows that while the serial baseline \hidrive declines sharply with agent density, \algname maintains robust throughput. The observed speedup, scaling from $227\times$ in sparse settings to $1073\times$ in dense traffic, is a result of multi-tiered parallelization, which overcomes computational bottlenecks in dense traffic, enabling real-time planning.

% \fourGridFigure{maincaption={\textbf{Architectural Scaling Analysis.} Multi-threading yields near-linear scaling, while SIMD vectorization maintains efficiency in high-density environments.},
%   label={ablation_architecture},
%   imgA={Ablation_MT_Absolute.png}, capA={MT Scaling (Throughput)},
%   imgB={Ablation_MT_Speedup.png},  capB={MT Scaling (Speedup Ratio)},
%   imgC={Ablation_SIMD_Absolute.png}, capC={SIMD Scaling (Throughput)},
%   imgD={Ablation_SIMD_Speedup.png},  capD={SIMD Scaling (Speedup Ratio)}
% }

\sideBySideFigure[t!]{
\textbf{Ablation: multi-threading.}
(Left) Edges/ms vs. traffic density.
(Right) Speedup over single-threaded (ST), showing near-linear scaling ($\sim$8$\times$).
}{fig::ablation_multi_thread}{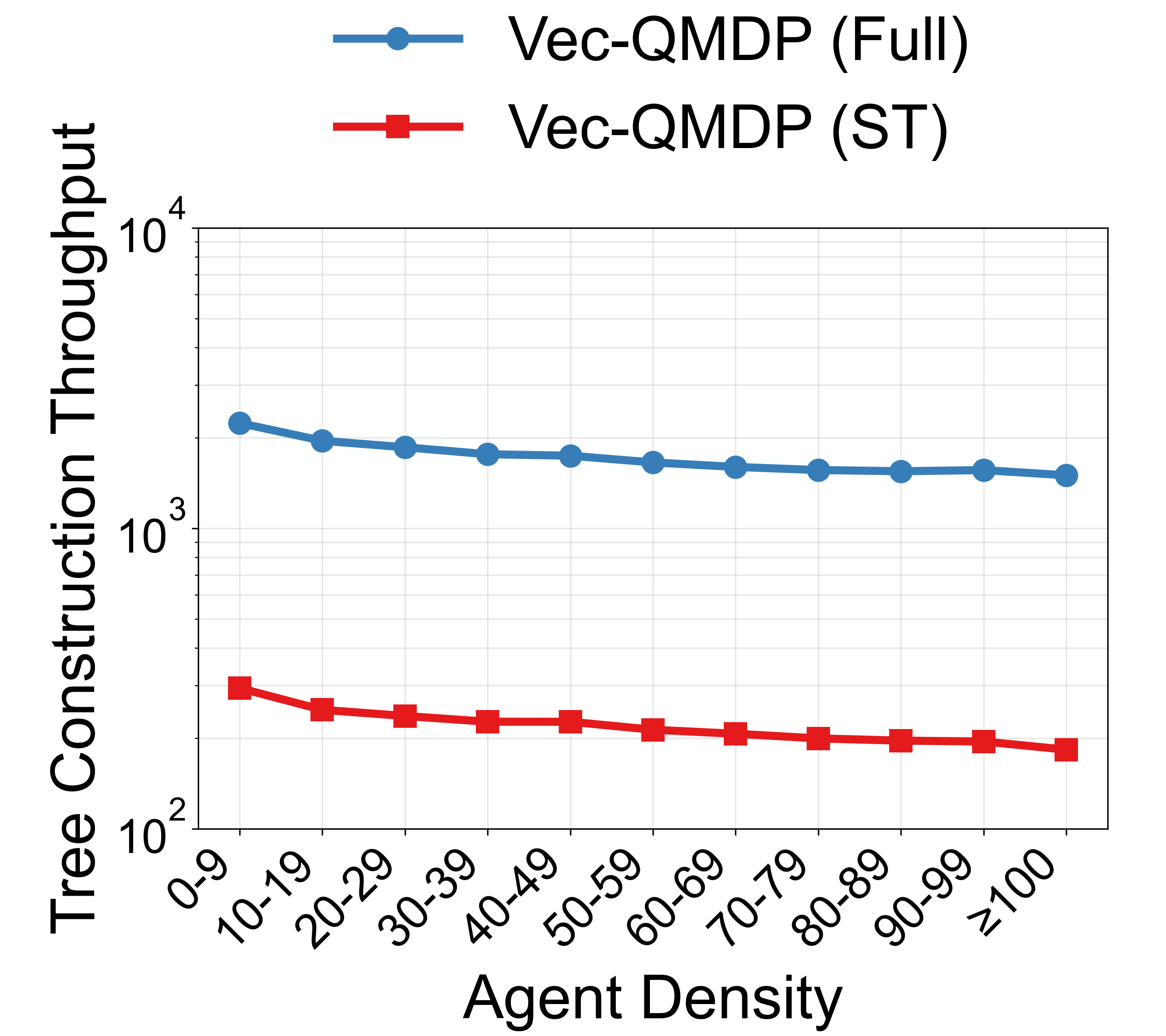}{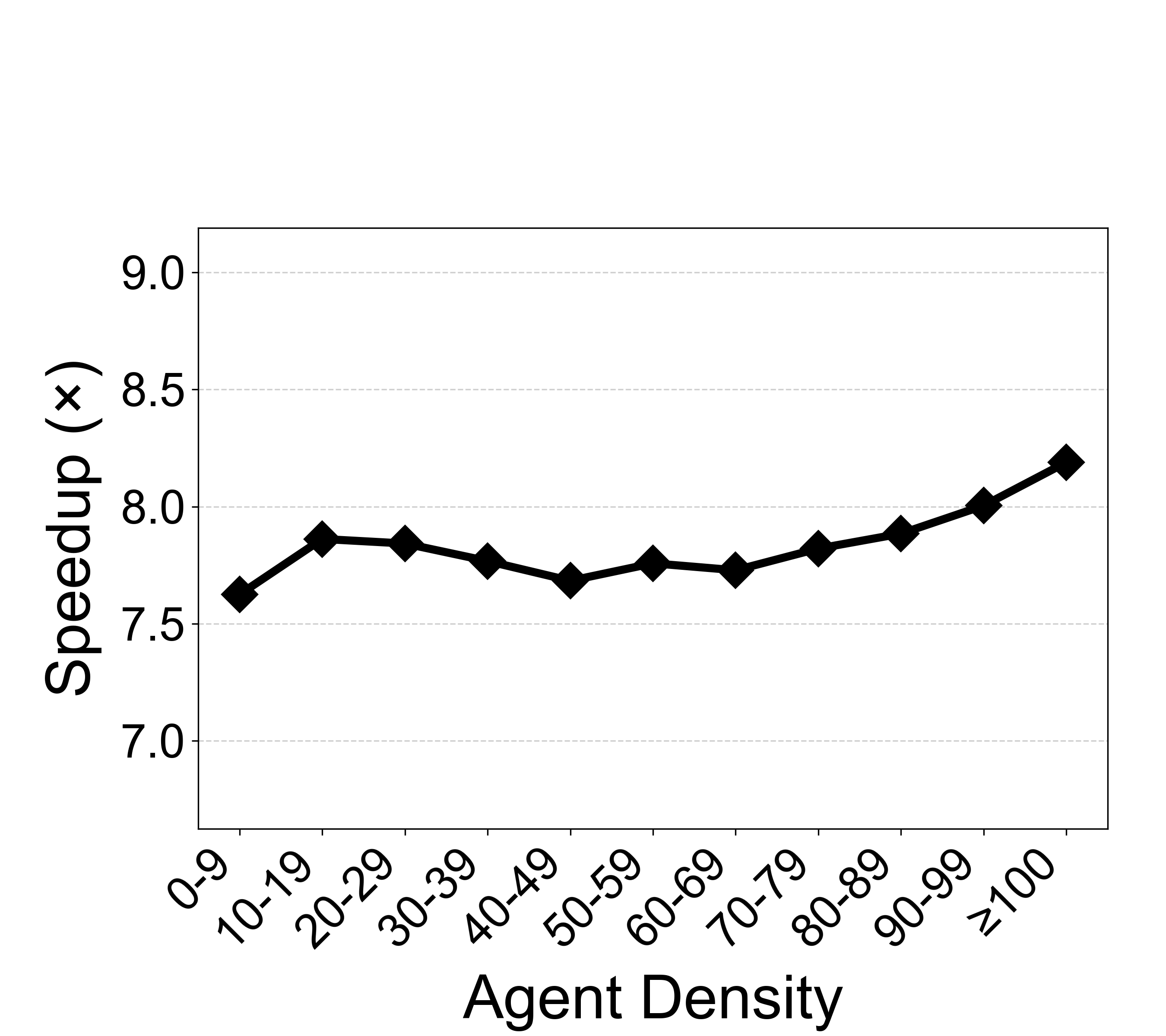}{Throughput (edges/$\mathrm{ms}$)}{Speedup (Full vs. ST)}

\subsection{Ablation Study on Multi-threading} \label{sec:ablation_mt}
We isolate the contribution of thread‑level parallelism by comparing our full multi‑threaded implementation ($M=8$ cores) with a single‑threaded variant (ST). Both use identical SIMD‑vectorized kernels and Data‑Oriented Design (DOD).

% \figref{fig::ablation_multi_thread} shows that tree‑construction throughput declines gradually for both configurations as agent density rises, reflecting the increased computational cost of dense traffic. However, the performance gap remains consistent across all densities, demonstrating that our strategy of distributing independent scenario sub‑trees across cores is largely independent of scene complexity. The resulting speedup stays near $8\times$, indicating nearly linear scaling with core count. This confirms that our lock‑free architecture eliminates synchronization overhead, enabling \algname’s search throughput to scale with available hardware resources.

\figref{fig::ablation_multi_thread} shows that while tree construction throughput declines gradually as agent density rises, the performance gap remains consistent across densities. The speedup remains near 8×, indicating nearly linear scaling with core count. This confirms that our lock‑free architecture eliminates synchronization overhead, enabling \algname’s search throughput to scale with available CPU cores.

\sideBySideFigure[t!]{
\textbf{Tree construction throughput.}
(Left) Edges/ms vs. traffic density.
(Right) Speedup over serial \hidrive\ (227$\times$--1073$\times$), increasing with density.
}{fig::throughput_comparison}{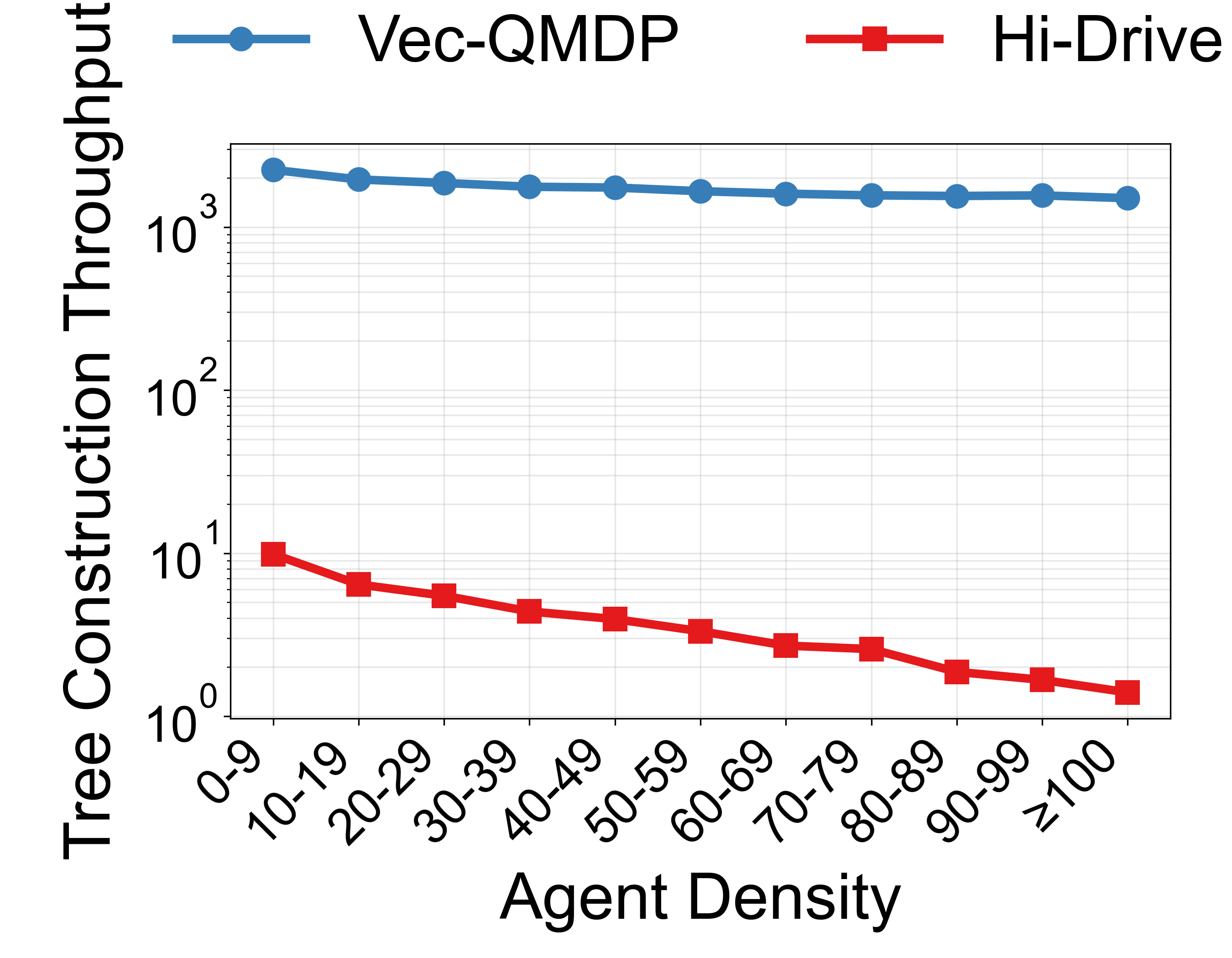}{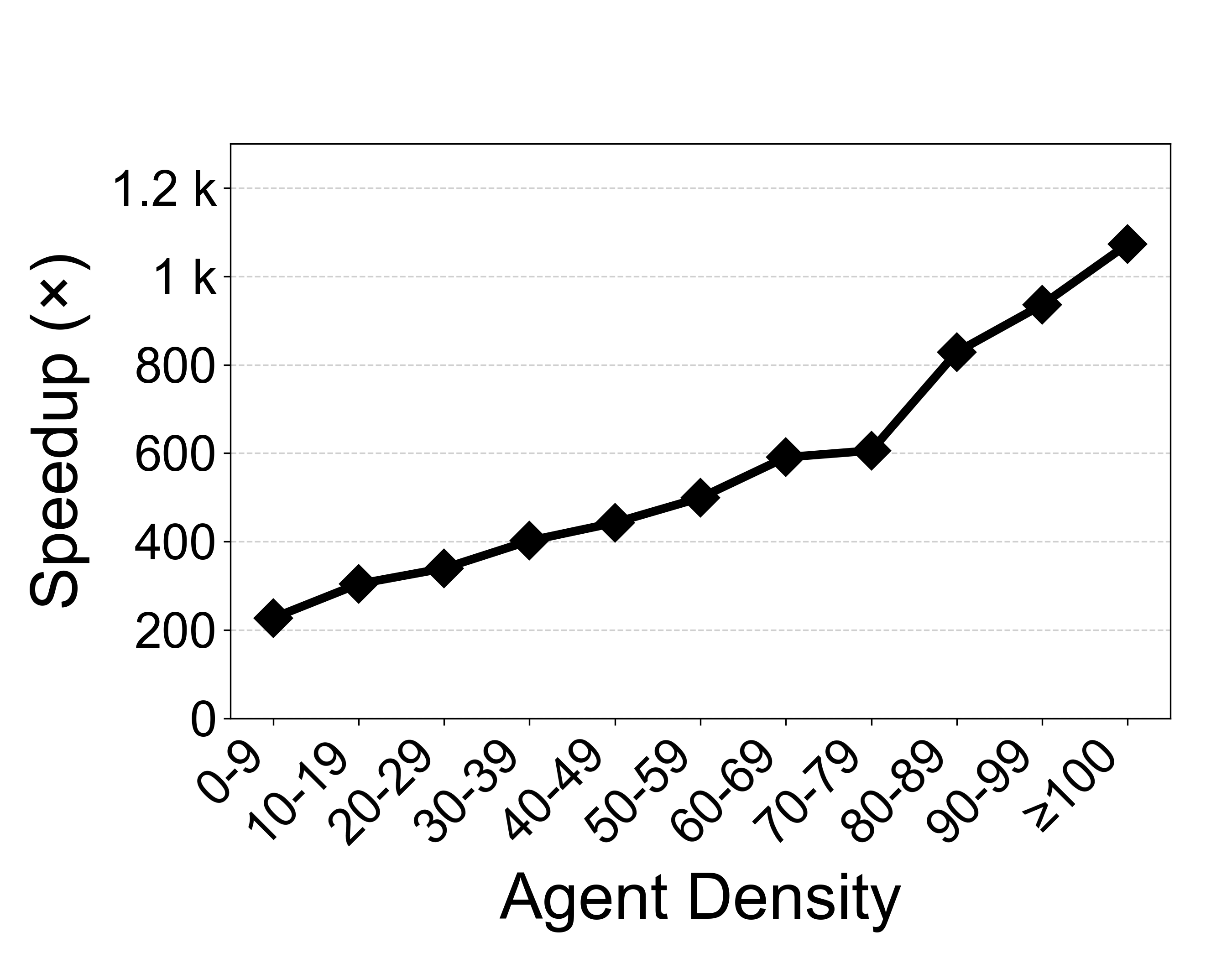}{Throughput (edges/$\mathrm{ms}$)}{Speedup over \hidrive}

\sideBySideFigure[t!]{
\textbf{Ablation: load-balancing UCB.}
(Left) Edges/ms vs. load imbalance.
(Right) Speedup over w/o LB, up to 1.24$\times$ in highly imbalanced scenes.
}{fig:load_balance}{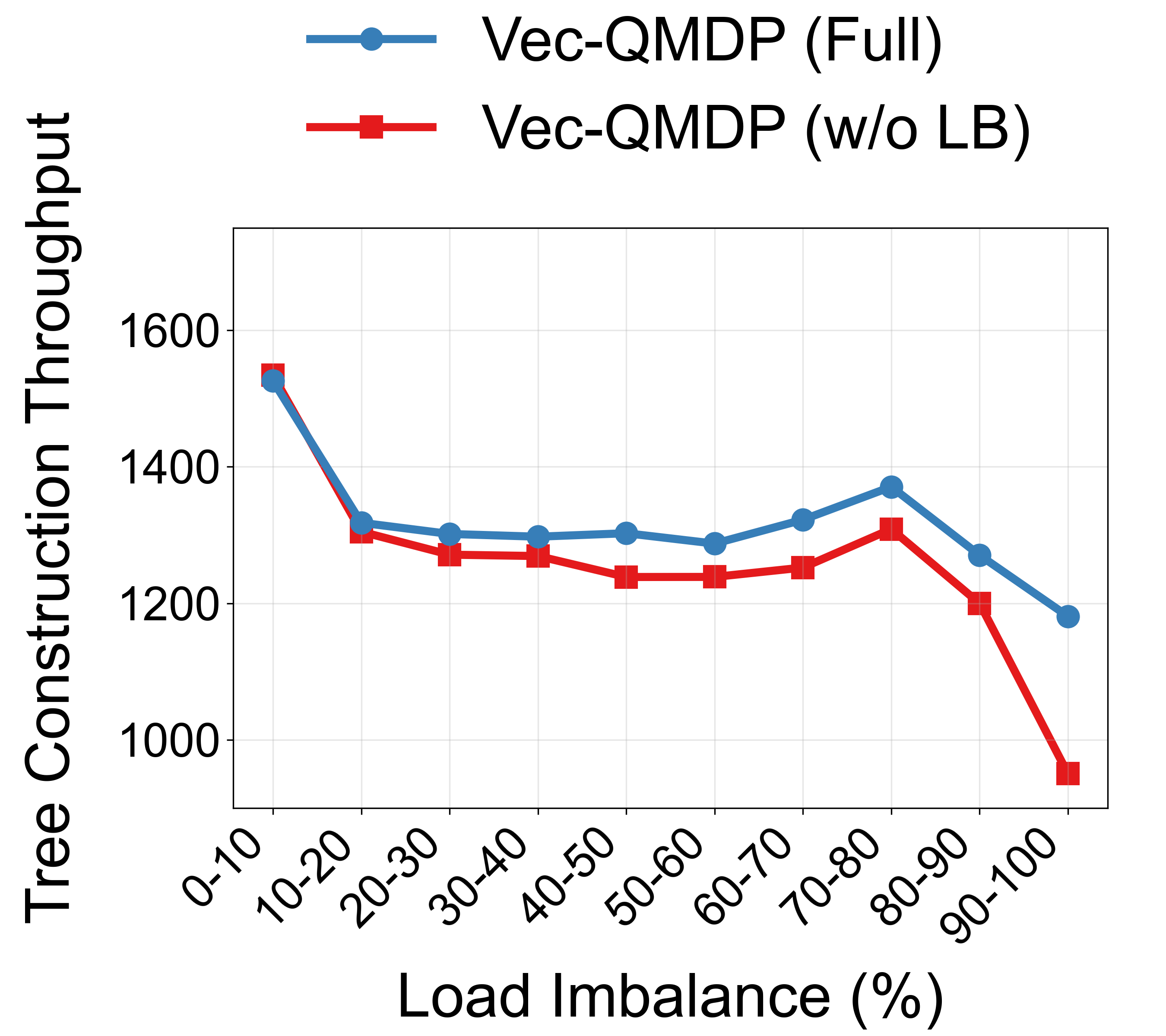}{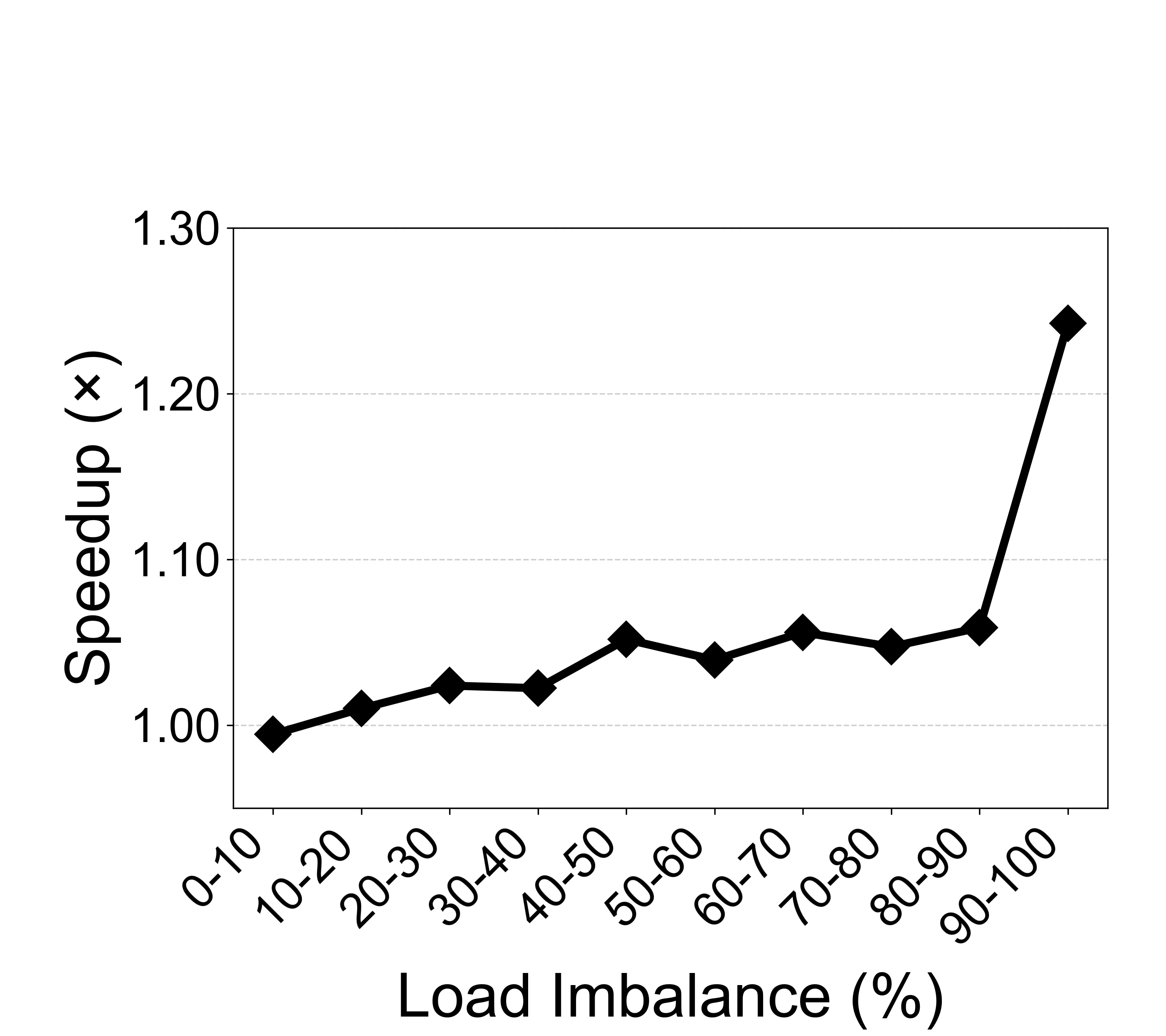}{Throughput (edges/$\mathrm{ms}$)}{Speedup (w/ LB vs. w/o LB)}

\begin{table}[t]
\centering
\small 
\caption{Ablation Study of Load Balancing on nuPlan.}
\label{table::lb_quality}
\renewcommand{\arraystretch}{0.95} 
\setlength{\tabcolsep}{10pt} 

\begin{tabular}{@{} l cccc @{}}
\toprule
& \multicolumn{2}{c}{\textbf{Val14} (NR)} & \multicolumn{2}{c}{\textbf{Test14-Hard} (NR)} \\
\cmidrule(lr){2-3} \cmidrule(lr){4-5}
\textbf{Method} & DS$\uparrow$ & SS$\uparrow$ & DS$\uparrow$ & SS$\uparrow$ \\
\midrule
w/o LB & 94.32 & 99.31 & 82.83 & 96.98 \\
w/ LB  & 94.36 & 99.36 & 82.84 & 96.76 \\
\bottomrule
% \multicolumn{5}{@{} l @{}}{
%     \scriptsize *DS: Driving Score.
% }
\end{tabular}
\end{table}

\subsection{Ablation Study on Load-Balancing UCB} \label{sec:load_balancing}
We ablate the proposed load-balancing UCB (Sec.~\ref{sec:lb_UCB}) by comparing the full planner (w/ LB) against a UCB-only variant (w/o LB).

\textbf{Load imbalance metric.} In the w/o-LB variant, we run standard UCB independently across scenario trees and obtain the expansion-depth ranges ${d_i}$. We define the depth difference as $\Delta d = \max_i d_i - \min_i d_i$, and measure \emph{load imbalance} as
\[
\text{Imbalance} \;=\; \frac{\#\{\text{simulations with } \Delta d \ge 1\}}{\#\{\text{total simulations}\}},
\]
i.e., the fraction of simulations whose expansion-depth ranges are misaligned across SIMD lanes. We compute this metric only from the w/o-LB run to capture the inherent depth divergence induced by plain UCB.

% Fig.~\ref{fig:load_balance} shows that the throughput gain (w/ LB over w/o LB) increases monotonically with the imbalance metric. When imbalance exceeds $90\%$, load-balancing UCB improves throughput by ${\sim}24\%$ by aligning selection depths (via $d_{\mathrm{ref}}$ and $\mathcal{D}(\cdot)$) and reducing SIMD lane idling during vectorized expansion/rollout.

\modified{
Fig.~\ref{fig:load_balance} shows that the throughput gain (w/ LB over w/o LB) increases monotonically with the imbalance metric. When imbalance exceeds $90\%$, load-balancing UCB improves throughput by ${\sim}24\%$ by aligning selection depths via $d_{\mathrm{ref}}$ and $\mathcal{D}(\cdot)$, thereby reducing SIMD lane idling during vectorized expansion and rollout. As shown in Table~\ref{table::lb_quality}, this throughput improvement does not materially affect planning quality: driving scores remain slightly higher with LB, while safety scores remain comparable on both Val14 and Test14-hard.
}

\subsection{Scalability Test on Planning Time} \label{sec:scalability}
We analyze how driving performance scales with planning time on the nuPlan benchmark, varying the time limit from 0ms to 100ms. \figref{fig:scalability_analysis} shows that both driving score and tree size stabilize within approximately 14ms. This rapid convergence indicates that the planner identifies high-quality driving trajectory almost immediately, resolving complex multi-agent interactions well within a real-time planning. The constructed tree size also stabilizes quickly, reaching a plateau that reflects the effective coverage of the belief space required for safe navigation.

These results demonstrate that our parallel POMDP planner efficiently scales performance with incremental CPU compute, and achieves near-optimal performance at 14ms, well below the conventional 100ms planning cycle.

\section{Conclusion}
We introduced \algname, a CPU-native framework for real-time POMDP planning that leverages hardware parallelism. By refactoring traditional pointer-based trees into vectorized structures using Data-Oriented Design (DOD), our approach combines multi-threading with SIMD vectorization. Through \textit{global vectorization}, we batch transition dynamics across independent scenario trees, while \textit{local vectorization} accelerates internal search components, such as collision checking. Additionally, load-balancing UCB aligns expansion-depth range across concurrent sub-trees to maximize SIMD lane utilization. Evaluation on the nuPlan benchmark demonstrates that \algname\ achieves $227\times$–$1073\times$ speedups in tree construction throughput, delivering state-of-the-art driving performance with millisecond-level planning times.

\modified{
The main limitation of \algname is its reliance on the QMDP approximation, which assumes that uncertainty is resolved after the initial action and therefore cannot fully capture deep active information gathering. Nevertheless, this assumption is well aligned with common autonomous-driving prediction pipelines, where a fixed set of future trajectories is generated before planning, while passive information gathering through future observations is still preserved. Moreover, when augmented at test time with an entropy-reduction reward following POMDP-lite~\cite{chen2016pomdp}, \algname achieves higher cumulative reward than the advanced POMDP solver DESPOT~\cite{ye_despot_2017} on RockSample(15,15)~\cite{smith2012heuristic} (16.824 vs. 13.356) with a 157$\times$ speedup (4.2 ms vs. 660.7 ms), suggesting that the practical loss of information-gathering capability can be partly mitigated. A second limitation is that SIMD efficiency depends on computational uniformity across scenario trees; highly heterogeneous dynamics may reduce lane utilization.

Future work will extend the proposed global and local vectorization principles beyond QMDP to more general online POMDP solvers such as DESPOT, enabling richer belief-dependent reasoning. Another promising direction is scaling \algname to large or high-dimensional action spaces, for example by combining vectorized POMDP search with learned macro-actions~\cite{Lee-RSS-21} for high-DOF robotic systems.
}

\insertFigure[t!][\linewidth]{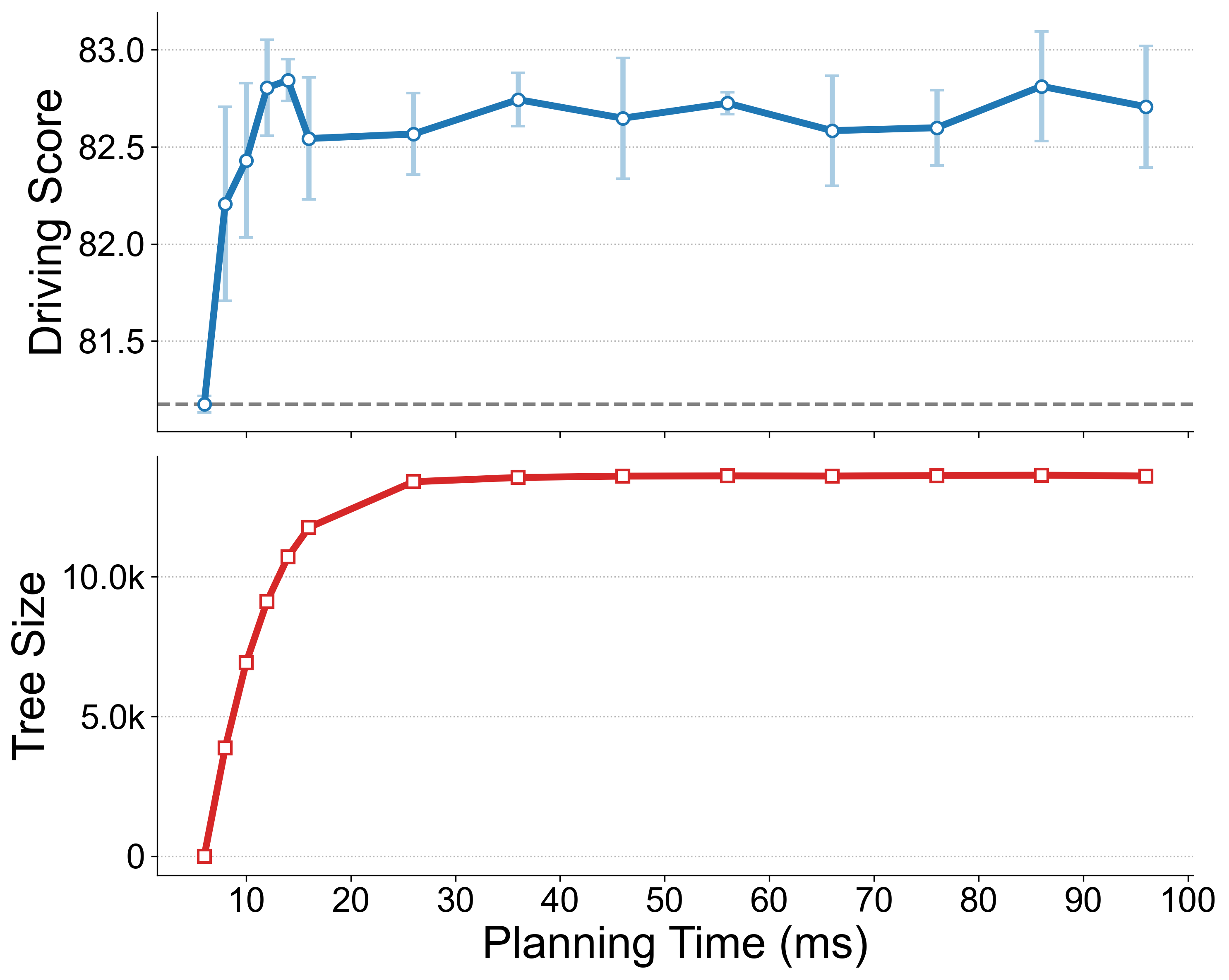}{
\textbf{Planning-time scalability.}
Driving score (top) and tree size (bottom) vs. planning time; both plateau by $\sim$14ms. Error bars: 95\% CI.
}{fig:scalability_analysis}

% Despite these results, the framework has limitations. First, the QMDP approximation assumes uncertainty resolves after the initial action, limiting its applicability to tasks requiring deep active information gathering. Second, while \textit{global vectorization} mitigates branch divergence, efficiency remains dependent on computational uniformity across scenario trees; highly heterogeneous transition dynamics may still lead to SIMD lane underutilization.

\section{Acknowledgments}
\modified{
This work is supported by New Generation Artificial Intelligence-National Science and Technology Major Project (No. 2025ZD0122901). This work was supported in part by the National Natural Science Foundation of China under Grant 62303304.
}

% \bibliographystyle{IEEEtran}
% \bibliography{IEEEfull, reference}

\bibliographystyle{plainnat}
\bibliography{reference}

\end{document}